\documentclass[runningheads]{llncs}
\usepackage{accv}
\usepackage{accvabbrv}
\usepackage{graphicx}
\usepackage{booktabs}
\usepackage{hyperref}
\usepackage{orcidlink}

\begin{document}
\title{Reference-Free Food Volume Estimation from Egocentric Stereo Video} 
\titlerunning{Reference-Free Food Volume Estimation from Egocentric Stereo Video}

\author{Vivan Kumar\orcidlink{0009-0007-4003-9291} \and Gautham Vinod\orcidlink{0000-0002-2488-2541} \and Fengqing Zhu\orcidlink{0000-0002-3863-3220}}
\authorrunning{V.~Kumar et al.}

\institute{Elmore Family School of Electrical and Computer Engineering, Purdue University, West Lafayette, IN 47907, USA \\
\email{kumar733@purdue.edu}}
\maketitle

\begin{abstract}
Current stereo vision-based approaches for food portion estimation are hindered by the user burden of using a mobile phone to capture multiple images or placing a reference object in the scene for scale calibration. As a step towards passive dietary intake monitoring, we introduce a method that uses stereo video from a wearable egocentric device to reconstruct food items in true physical scale to accurately estimate volume. By utilizing FoundationStereo for zero-shot stereo matching, SAM 2 for food segmentation, and TSDF fusion over a set of keyframes, our method yields a visually and metrically accurate 3D mesh of any plated food item from a single stereo recording. We leverage the unique offerings of wearables, such as a factory-calibrated stereo baseline and camera poses estimated on-device via visual-inertial odometry (VIO), to perform fully automatic reconstruction without a scale reference object in the scene. We show strong initial volume estimation results across five recorded food items, achieving a mean absolute percentage error of 9.1\%.
\end{abstract}

\section{Introduction}
Image-based methods have introduced a new paradigm for dietary monitoring, enabling food portion estimation using one or a few images captured on consumer devices such as mobile phones. However, these methods are constrained by the loss of 3D information in monocular images, requiring methods such as monocular depth estimation and reference objects placed in the scene to estimate the true physical scale of food items \cite{graikosmono,metafood2024}. Alternatively, stereo images more naturally mimic how the human eyes perceive depth and provide a significantly more accurate cue for estimating true scale through pixel disparity; however, existing methods built around stereo images maintain the user burden of manually capturing images on a mobile phone and placing a reference object in the scene  \cite{dehaismulti}. These methods are also more likely to fail on glossy or textureless food items due to the fragility of traditional stereo matching methods.

With the rise of wearable consumer devices equipped with stereo cameras and rich sensor suites \cite{snapspecs}, food portion estimation methods are transitioning towards approaches based on egocentric vision \cite{vinodstereo, egocal, aim2}. Egocentric videos from wearables can capture natural eating scenes, potentially unlocking passive dietary monitoring; however, this shift will only prove to be useful if data provided by wearables (such as known camera calibrations and estimated poses) is uniquely harnessed to estimate true metric scale of food items without a substantial user burden. To this end, we propose a method that leverages stereo videos captured on the egocentric wearable Project Aria Gen 2 glasses \cite{ariagen2} to estimate the volume of plated food items without any reference objects required in the scene. 

Our method runs end-to-end on recorded videos where the wearer ``scans'' a plated food item, utilizing FoundationStereo \cite{foundationstereo} for zero-shot metric-scale depth estimation and SAM 2 \cite{sam2} for food mask segmentation over a set of keyframes. A 3D mesh of the food item is extracted using truncated signed distance function (TSDF) fusion \cite{tsdf} over a voxel grid representing the shared global coordinate frame; this mesh is metric by construction and requires no further scaling. Our video recording protocol has a relatively high user burden due to the need for a deliberate scanning movement around the plate by the wearer, but the strong volume estimation results from our experiments demonstrate the viability of our approach as a step towards passive dietary monitoring. The contributions of this paper can be summarized as follows:
\begin{itemize}
\item{A generalizable method to extract metric-scale 3D meshes of food items to estimate volume from egocentric stereo videos captured on the Project Aria Gen 2 wearable glasses.}
\item{A pipeline that runs end-to-end on a single recording file following our recording protocol, automatically deriving a watertight mesh even though the contact region between the food and the plate is never observed by the wearer.}
\item{Experiments demonstrating our accurate volume estimations and reconstructions, along with a comparison demonstrating the necessity of stereo depth estimation over monocular metric depth estimation in this approach.}
\end{itemize}

\begin{figure}[tb]
  \centering
  \includegraphics[height=3cm]{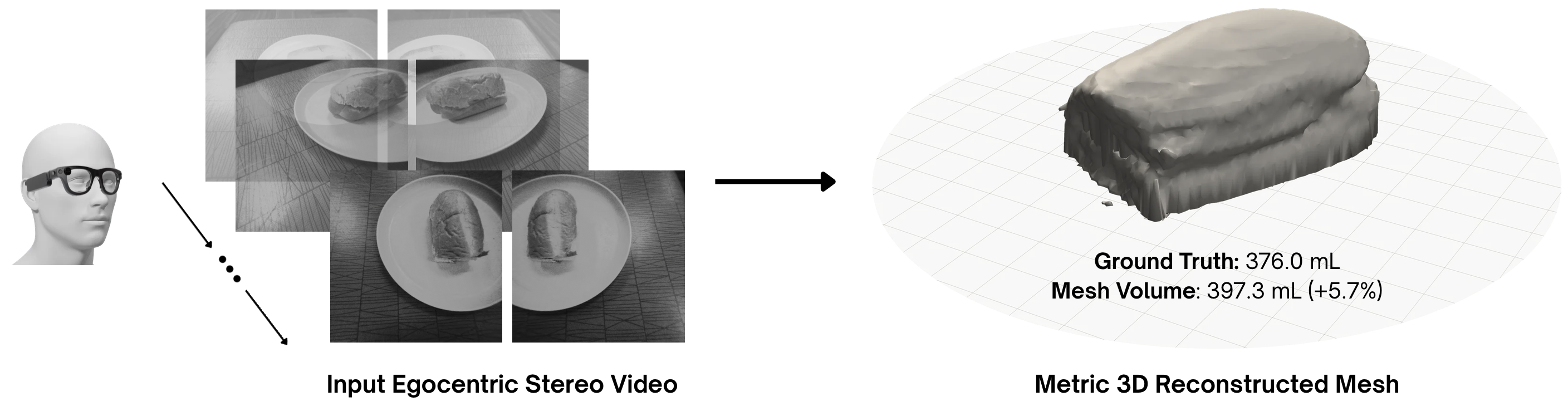}
  \caption{Our method builds a metrically accurate watertight mesh of a plated food item from two full arcs of head motion around the plate. The true physical scale of the item is recovered entirely from the stereo cameras’ calibration and on-device pose estimation. No reference object or user-provided dimension is used at any stage.
  }
  \label{fig:teaser}
\end{figure}

\section{Method}
\begin{figure}[tb]
  \centering
  \includegraphics[height=7.3cm]{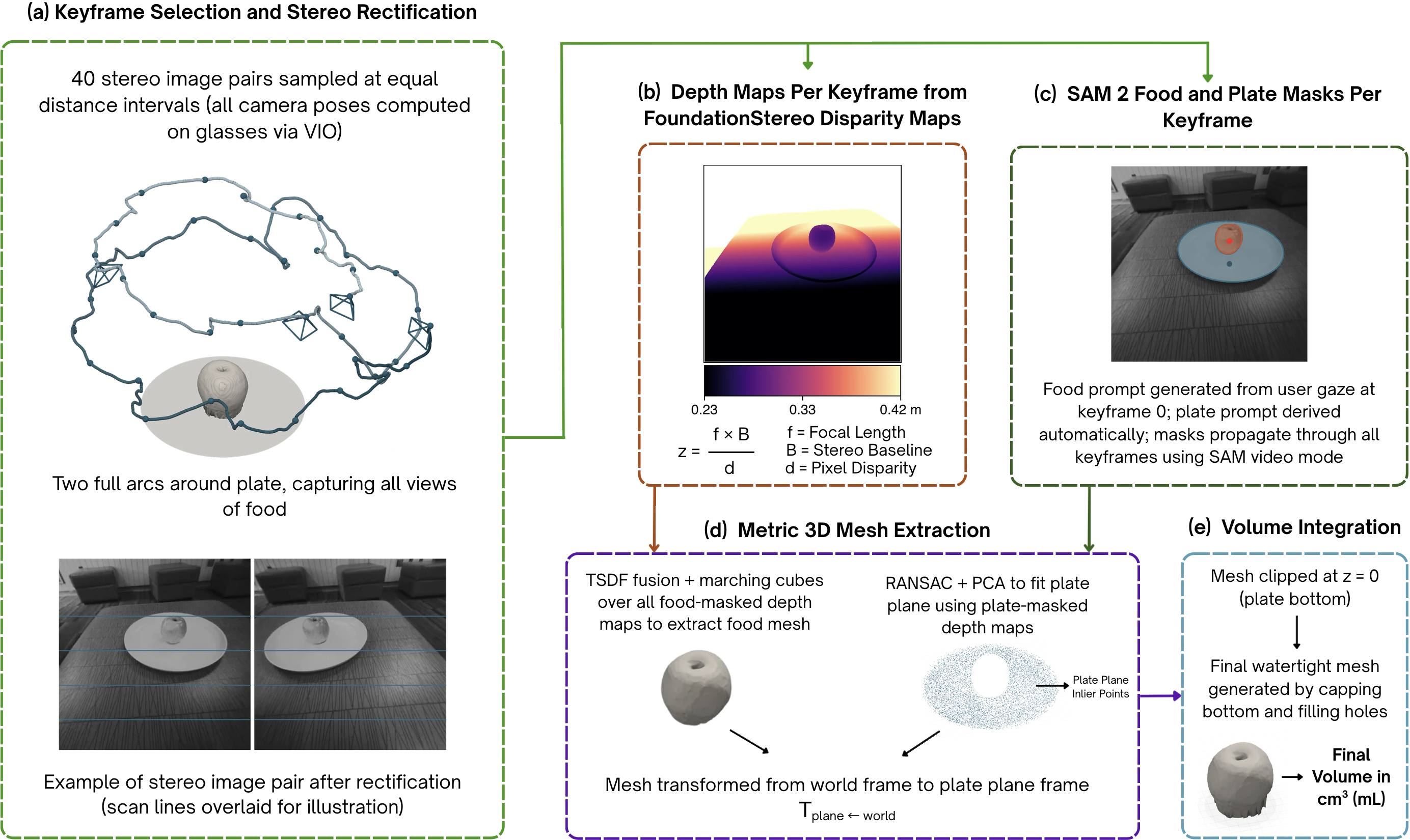}
  \caption{\textbf{Method overview.} (a) 40 rectified stereo keyframes spaced evenly across the device trajectory are used as inputs to (b) FoundationStereo. Metric depth maps are computed from the generated dense disparity maps using the stereo baseline and camera intrinsics. (c) Food and plate masks are generated using SAM 2 by leveraging the wearer’s gaze on the food in the first keyframe. (d) The depth maps and food masks yield a mesh using traditional TSDF fusion, whose contact region with the plate is unobserved in the camera views, guaranteeing a hole; (e) this hole is filled using the plate plane fitted in the previous stage by clipping the mesh at the plane’s surface. All holes are filled via fan triangulation, and the final watertight mesh is integrated to estimate the food’s volume.
  }
  \label{fig:pipeline}
\end{figure}

The input to our method is a VRS file that bundles all data streams recorded by the Aria glasses. We rely on the following streams: monochrome video from the front stereo camera pair, left camera poses (where a pose is a $4 \times 4$ transformation matrix $T_{world \leftarrow cam}$ that transforms points in the left camera's frame to the world frame) computed on-device using VIO, and the user's gaze pitch and yaw computed on-device using the eye-tracking cameras. This multimodal data captured in our 30-second recordings enables our five-stage pipeline illustrated in \cref{fig:pipeline}, which runs end-to-end without manual supervision. 

\subsection{Keyframe Selection and Stereo Depth}
We use the left camera's poses in each captured frame to accumulate the distance traveled by the glasses in the world frame, sampling 40 keyframes from the synchronized left and right cameras uniformly along this path. Sampling based on distance traveled allows us to capture diverse views of the food, which is crucial for TSDF fusion. Both cameras contain fisheye lenses, so all keyframe pairs are undistorted and rectified using the known fisheye calibration to obey the pinhole model. Each rectified keyframe pair is passed to FoundationStereo with the ViT-L backbone, which performs learned stereo matching and outputs dense disparity maps \cite{foundationstereo}.

We utilize FoundationStereo because of its strong zero-shot generalizability compared to traditional stereo matching methods such as Semi-Global Block Matching (SGBM) which would be more likely to fail on reflective surfaces such as glass plates and glossy food items \cite{sgbm}. Each resulting disparity map is used to compute a metric-scale depth map using the Aria glasses' known stereo camera baseline and the focal length in the rectified images.

\subsection{Food and Plate Segmentation}
The Aria glasses use two eye-tracking cameras to estimate the wearer's gaze direction and the shared depth of the point in the scene where both eyes focus \cite{ariagen2}. As a step towards fully passive food portion estimation, we leverage the wearer's gaze as a point prompt for SAM 2 \cite{sam2} to segment the food item across all keyframes. Crucially, because the depth maps are computed for images from the left camera of each stereo pair, all masks are also computed on the left images. 

We begin by naively projecting the wearer's 3D eye gaze point in the first keyframe into a 2D point in the image, as each recording begins with the wearer focusing their gaze on the food item. This approach could be strengthened by future approaches that first detect eating episodes before using gaze data for segmentation prompts. Additionally, we use a simple algorithm to automatically derive a segmentation prompt for the plate in the scene using the original 2D gaze point and first food mask. We utilize SAM 2's ability to handle videos via its memory system \cite{sam2} to derive boolean food and plate masks for all 40 keyframes using the single $(x, y)$-coordinate gaze prompt from the first keyframe. 

\subsection{Metric 3D Mesh Extraction}
We utilize the Open3D library \cite{open3d} to define a grid of 2.5 mm voxels in the world frame for TSDF fusion, where each voxel (automatically truncated to a small region around the surface during fusion) stores the signed distance from its center to the nearest observed surface. We do not store RGB colors in the voxels since they are unnecessary for volume computation, but colors can be fused across views using the RGB camera intrinsics to produce realistic colored food meshes, which we do not evaluate here. Using the camera intrinsics and poses, we perform TSDF fusion over all 40 food-masked depth maps, which averages out minor depth discrepancies across the multiple views and yields an implicit representation of the food surface in the voxel grid.

We employ Open3D's marching cubes algorithm method \cite{marching} over the voxel grid and recover a food mesh, which is metrically accurate and requires no further scaling steps. In order to fill the hole in the mesh caused by the contact region between the food and plate not being observed in any view, we first fit the plate surface plane in the world frame using RANSAC \cite{ransac} over a subset of the plate's 3D points. We center the plane inliers (plate points that lie on the fitted plane within a small distance threshold) and apply PCA to them to recover a final accurate plate plane estimate. Finally, we transform each vertex in the food mesh from the world frame into a newly defined plate coordinate frame using the fitted plane to aid in hole-filling in the final stage.

\subsection{Watertightness and Volume Integration}
We define a new coordinate frame on the plate, where the plate's surface is the plane $z = 0$, and positive $z$ points upwards. Any triangle in the produced mesh that straddles the $z = 0$ plane is clipped so that the mesh lies entirely above the plate plane. We then classify each of the boundary loops (series of edges forming holes) in the mesh: a contact loop is formed by the unobserved region where the food meets the plate's surface, and a pinhole loop is a small artifact that may sometimes be produced during the marching cubes step. Both types of loops are closed by defining a new vertex in their center and drawing edges from that vertex to every other vertex in the loop to form new triangles. This produces a final watertight mesh of the food item that is integrated immediately without any scaling steps to yield the final computed volume.

\section{Experiments}
\subsection{Dataset}
To evaluate our approach, we record five 30-second videos on the Aria glasses: in each recording, the wearer walks two complete circles around a plated food item: one observing it from a high elevation above the table ($\sim60^\circ$) and the other from a lower elevation ($\sim30^\circ$) to capture the food's outer surface from multiple views for accurate TSDF fusion. All necessary video and sensor streams are automatically recorded in the VRS file for each recording. All frames are recorded at a resolution of 512 $\times$ 512 pixels, but our method is not tied to the image resolution used.

Ground truth volumes for all food items (apple, honey bun, pizza slice, sandwich, and sub sandwich) are obtained using water displacement. Because all selected food items excluding the apple are porous, they are first covered in a water-resistant plastic wrap to create an airtight seal during measurement; they are also buoyant and are thus submerged into the water using thin bamboo skewers. The volume occupied by the plastic wrap and bamboo skewers in the water is negligible for our evaluation.

\subsection{Volume Estimation Results}
We evaluate our method quantitatively using the mean absolute percentage error (MAPE) between the ground truth measurements and estimated volumes. Our results are presented in \cref{tab:results}: our method achieves a MAPE of 9.1\% across all five tested food items. Considering that the food items in our dataset are relatively low in volume, and that every estimate lies within 13\% of the ground truth, our results demonstrate our method's strong ability to measure food portion volumes without any learned food volume priors or scale estimation steps. We have not isolated the cause of the underestimations or overestimations of volume, due to the varying geometries of the food items used and the limited dataset.

\begin{figure}[tb]
  \centering
  \includegraphics[height=3.2cm]{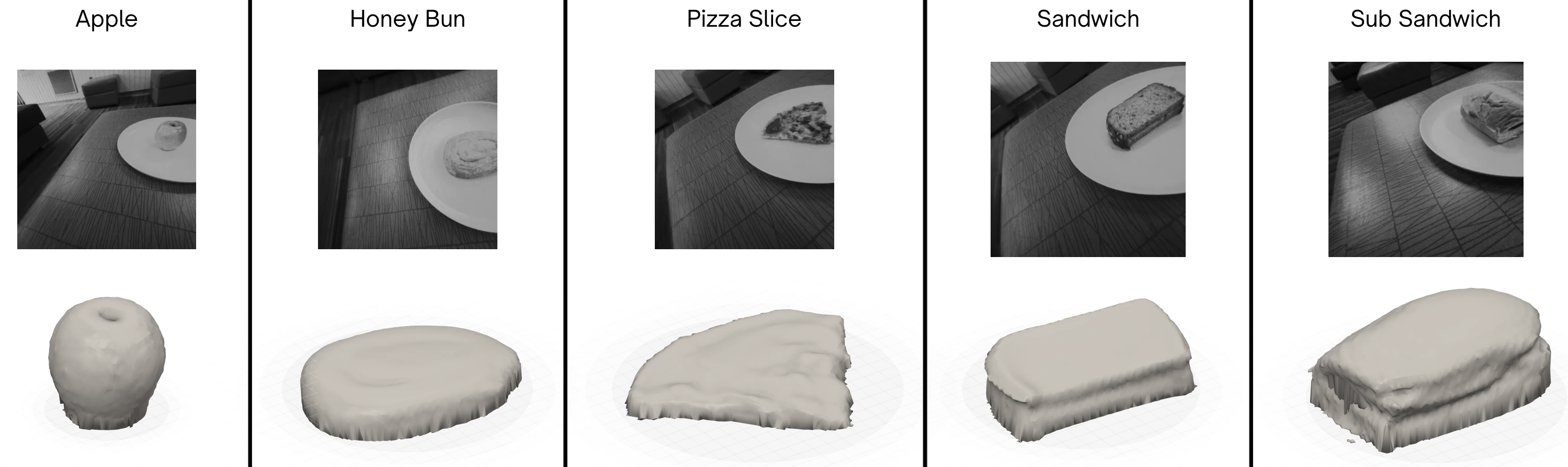}
  \caption{\textbf{Qualitative results on food reconstruction.} Our method leverages TSDF fusion to produce meshes that capture fine geometric details, such as the cavity on the apple (first column) and the irregular toppings on the pizza slice (third column).
  }
  \label{fig:gallery}
\end{figure}

In \cref{fig:gallery}, we visualize the reconstructed 3D meshes of all food items. TSDF fusion across 40 well-spaced keyframes yields meshes that are not only in true physical scale but also accurately capture the surface geometry of each food item. However, the clipping and hole-filling steps sometimes produce triangles near the bottom of a mesh that appear to ``snap'' down to the surface plane, as observed in the reconstructed apple in \cref{fig:gallery}. These visual artifacts enclose an extremely small amount of volume that does not meaningfully skew our total volume estimates.

\begin{table}[tb]
  \caption{Quantitative results of volume estimation against ground truth.}
  \label{tab:results}
  \centering
  \setlength{\tabcolsep}{10pt}
  \begin{tabular}{lrrr}
    \toprule
    Item & Ground Truth (mL) & Estimated\ (mL) & Error\ (\%) \\
    \midrule
    Apple         & 177.0 & 156.2 & $-$11.8 \\
    Honey Bun     & 219.0 & 190.9 & $-$12.8 \\
    Pizza Slice   & 141.0 & 125.2 & $-$11.2 \\
    Sandwich      & 189.0 & 196.6 & $+$4.0 \\
    Sub Sandwich  & 376.0 & 397.3 & $+$5.7 \\
    \midrule
    \multicolumn{3}{l}{MAPE} & \textbf{9.1\%} \\
    \bottomrule
  \end{tabular}
\end{table}

\subsection{Stereo vs. Monocular Depth}
To demonstrate the importance of using stereo image pairs for depth estimation in our method, we run the entire pipeline for each of the five food items, substituting only the depth maps for each keyframe with those produced by Depth Anything V2 \cite{depthanything} (the \textit{Depth-Anything-V2-Metric-Indoor-Large-hf} checkpoint), a monocular metric depth estimation model. No valid mesh is produced in these runs due to overestimated depth values at this close viewing range and inconsistency across different views, resulting in TSDF and marching cubes failing. Specifically, when point clouds of any given food item from different keyframes are produced by backprojecting the food pixels into the 3D world frame using their monocular metric depth values and the camera intrinsics, they do not overlap as expected. 

\begin{figure}[tb]
  \centering
  \includegraphics[height=4cm]{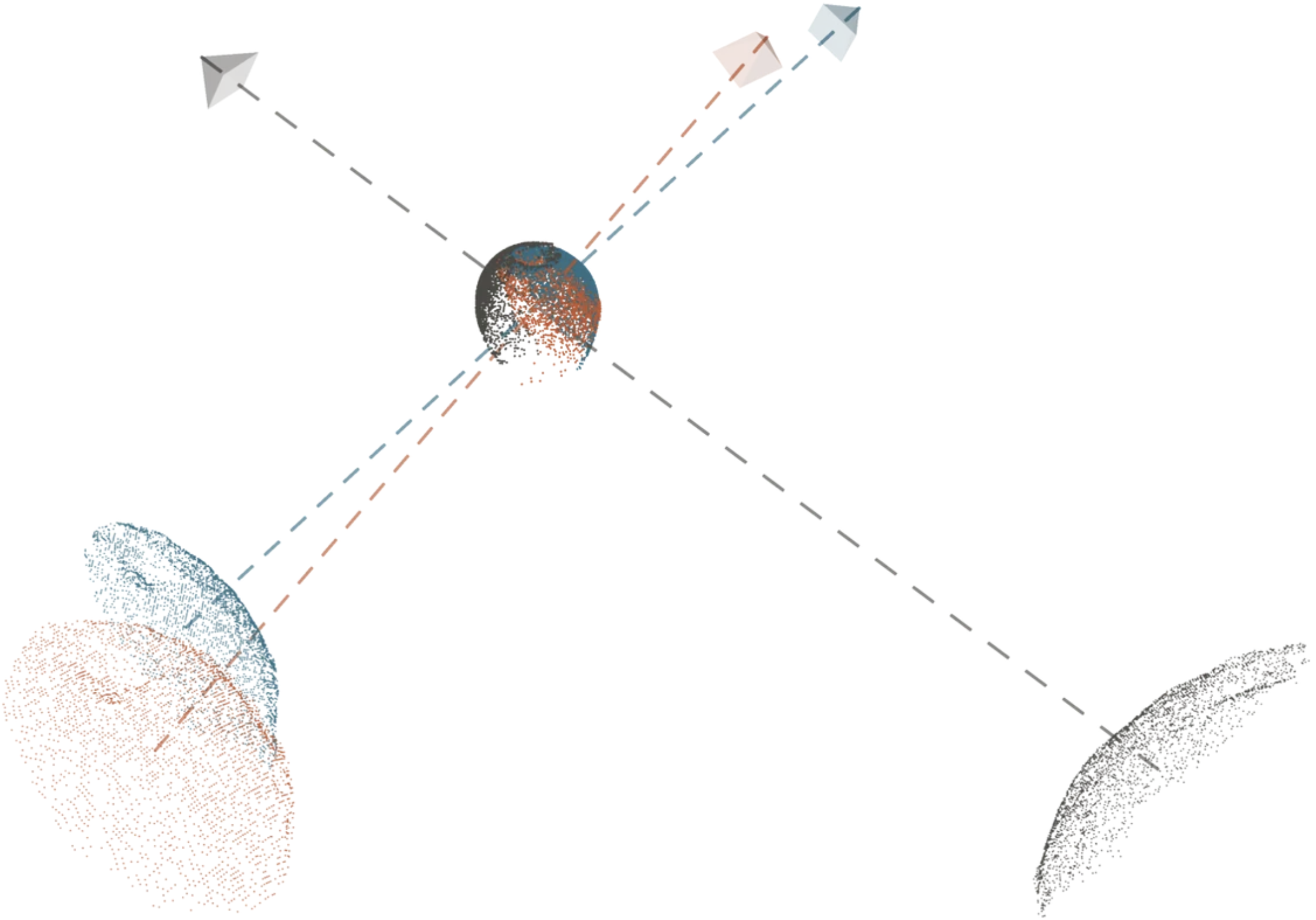}
  \caption{\textbf{Effect of replacing stereo with monocular depth estimation.} Point clouds of the apple captured from three different keyframes (separated by color) produced by backprojecting 2D pixels into the shared 3D world frame using the same poses. The point clouds produced using the FoundationStereo-based depth maps intersect and form a coherent shell of an apple; those produced using the monocular model’s depth maps do not intersect, resulting in failed TSDF fusion.
  }
  \label{fig:mono}
\end{figure}

An example of this failure is visualized in \cref{fig:mono}, where the monocular point clouds for each keyframe are 2--3$\times$ too far along the camera's viewing ray, resulting in point clouds that disagree on where the apple's points lie in the world frame. Because FoundationStereo performs learned stereo matching and computes disparity rather than inferring depth directly from learned priors, the point clouds generated using FoundationStereo depths are highly consistent across keyframes, leading to accurate TSDF fusion and the ability for our method to generalize to unseen food items and eating scenes. 

\section{Conclusion}
We present a method for reconstructing metrically accurate meshes of food items from egocentric stereo videos for volume estimation. By applying TSDF fusion over several keyframes with metric depth maps, we recover the true scale of food items with no reference object in the scene. On our dataset of five videos, we achieve a MAPE of 9.1\%, highlighting the potential for further approaches that leverage stereo videos and rich sensor data from emerging egocentric wearables for passive food portion estimation. \\
\textbf{Limitations.} Our method is built using a specific recording protocol that does not accurately represent dynamic eating episodes, where the wearer's hands and utensils occlude the food. Additionally, our protocol requires the wearer to deliberately capture all views of the food item in one pass, ensuring that neither the food nor the plate move in the scene during the scanning motion. A natural direction that emerges for further work is the tracking and reconstruction of food items as the wearer and the food move naturally throughout eating episodes. 

\bibliographystyle{splncs04}
\bibliography{main}
\end{document}